\documentclass[11pt]{article}

\usepackage{UF_FRED_paper_style}
\usepackage{float}
\DeclareMathOperator*{\argmax}{arg\,max}

\usepackage{lipsum}  

\title{CQ-VQA: Visual Question Answering on Categorized Questions}

\author{Aakansha Mishra\\
    \href{ak.kkb@iitg.ac.in}{\texttt{ak.kkb@iitg.ac.in}} 
\and Ashish Anand\\
    \href{anand.ashish@iitg.ac.in}{\texttt{anand.ashish@iitg.ac.in}} 
\and Prithwijit Guha\\
    \href{pguha@iitg.ac.in  }{\texttt{pguha@iitg.ac.in}}
    }
    
\date{}

\begin{document}
{\setstretch{.8}
\maketitle
\begin{abstract}
This paper proposes \textit{CQ-VQA}, a novel 2-level hierarchical but end-to-end model to solve the task of visual question answering (VQA). The first level of CQ-VQA, referred to as question categorizer (QC), classifies questions to reduce the potential answer search space. The QC uses attended and fused features of the input question and image. The second level, referred to as answer predictor (AP), comprises of a set of distinct classifiers corresponding to each question category. Depending on the question category predicted by QC, only one of the classifiers of AP remains active. The loss functions of QC and AP are aggregated together to make it an end-to-end model. The proposed model (CQ-VQA) is evaluated on the TDIUC dataset and is benchmarked against state-of-the-art approaches. Results indicate competitive or better performance of CQ-VQA.

\end{abstract}
}


\section{Introduction}

The objective of a Visual Question Answering (VQA) system \cite{antol2015vqa,Agrawal2017} is to generate a natural language answer to a natural language question asked about a given image. VQA has gained wide attention for several reasons. First, it has got many real-life applications involving scene interpretation for assistance to visually impaired persons, interactive robotic systems etc.. Second, it is a challenging AI problem as it requires a simultaneous understanding of two modalities -- image and text, and reasoning over the relations among the modalities. This wide attention has naturally led to the development of a plethora of methods.

The early approaches of VQA primarily focused on feature fusion of two modalities, where image- and text-based features are fused using simple techniques like addition, concatenation, or element-wise products \cite{antol2015vqa,zhou2015simple}. Later, improved feature fusion mechanisms such as bilinear pooling \cite{gao2016compact} and its variants MCB \cite{gao2016compact}, MFB \cite{yu2017multi}, MLB \cite{kim2016hadamard} and MUTAN \cite{ben2017mutan} were proposed.

Another class of methods focus on identifying `relevant' image regions for answering to the given question. Attention-based methods \cite{shih2016look,yang2016stacked,kazemi2017show,lu2016hierarchical,xu2016ask} fall into this category. These methods aim to assign higher weights (attention scores) to the image regions pertinent to answer the given question while providing relatively negligible attention to other regions. It is noteworthy to mention that such methods do fuse features of the different modalities. However, performance improvement significantly depends on the extent of information obtained by exploiting attention in different modalities. For example, studies in \cite{lu2016hierarchical,nguyen2018improved} have shown that along with question guided attention on image, attention from image to questions allow better information flow and interaction between the two modalities, resulting in improved performance.

\begin{figure}[t]
\centering
\includegraphics[width=\textwidth,height=8cm,keepaspectratio]{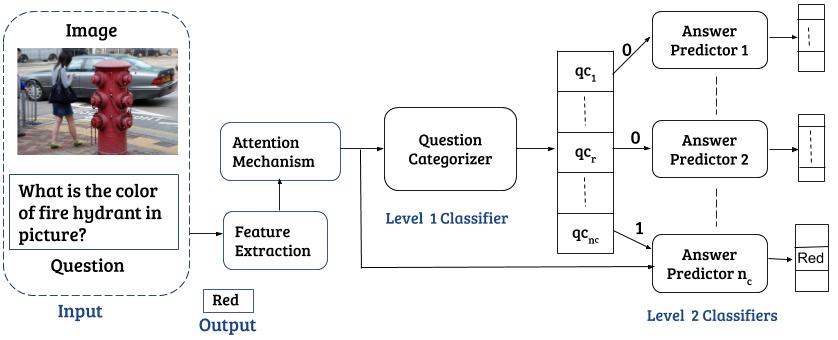}
\caption{An overview of the proposed framework of CQ-VQA. features extracted from input question about an image are fused through an attention mechanism. The hierarchical structure of CQ-VQA first categorizes the input question (level 1 classifier) and accordingly selects an answer predictor for identifying the output answer.}
\label{fig:introFig}
\end{figure}

This paper proposes a hierarchical model, referred to as \textit{CQ-VQA}. CQ-VQA hierarchically solves the VQA task by breaking it into two sub-problems. Figure~\ref{fig:introFig} illustrates the motivation and working principle of the CQ-VQA. As illustrated, a question ``\textit{What is the color of fire-hydrant in picture?}'' is asked about the given image. As a human, we immediately understand that the question is about the color of an object, and the answer must be one of the colors. CQ-VQA mimics this intuition in a 2-level hierarchical classification model. At the first level, a single classifier identifies the question category based on the fused features of the given question and image. Based on this classification, the CQ-VQA model sends fused features to one of the classifiers of second level. The second level contains a set of distinct classifiers, one for each question category and output of each classifier is set of answers belongs to that category. In contrast to the existing VQA models, where they need to explore the entire search space of answers, CQ-VQA focuses on smaller answer search spaces in the final stage of classification.

The performance of CQ-VQA is evaluated on the TDIUC dataset \cite{kafle2017analysis} containing $12$ explicitly defined question categories. The experimental results on this dataset have shown competitive or better performance of CQ-VQA compared to state-of-the-art models. The primary contributions of this work are as follows.

\begin{itemize}
\item A novel hierarchical model for decomposing the VQA task into two sub-problems -- question categorization and answer prediction.
\item End-to-end model training model by combining the two loss functions of the two sub-problems.
\item Comprehensive overall and question category-wise performance analysis and comparison with state-of-art VQA models.
\end{itemize}

The rest of the paper is organized as follows. A brief review of VQA literature is presented in Section~\ref{sec:relWork}. Section~\ref{sec:propWork} discusses the necessary details of the proposed approach. The experimental results are presented and discussed in Sections~\ref{sec:expRes} and ~\ref{sec:resDisc}. Finally, we conclude in Section~\ref{sec:conc} and sketch the extensions of the present proposal.

\section{Related Work}
\label{sec:relWork}

Existing works in VQA can be broadly divided into three categories. These are (a) feature fusion based approaches, (b) attention based methods, (c) reasoning based techniques. This proposal uses attention models for visual and question feature fusion. Accordingly, the existing works in the first two categories are briefly reviewed next.

\subsection{VQA: Feature Fusion}
\label{subsec:featFuse}

These approaches project both visual and question embeddings to a common space to predict the answer. The embeddings of the visual modality are obtained using pre-trained CNNs. These networks are learned from large image data sets dealing with different classification problems \cite{szegedy2015going,he2016deep,krizhevsky2012imagenet}. The questions are represented in two ways. The first class of approaches have used Bag-of-Words (BoW) representations for questions \cite{antol2015vqa,zhou2015simple,jabri2016revisiting}. The second group of methods represent questions as sequences of word2vec embeddings  \cite{mikolov2013efficient,pennington2014glove}. These embedding sequences are further input to pre-trained Recurrent Neural Networks (RNNs) for obtaining question embeddings \cite{mikolov2013efficient,pennington2014glove}. A third group of approaches represent questions using pre-trained CNN features \cite{ma2016learning,wang2018learning}. However, most existing works use the second method involving word2vec embedding sequence and RNN.   

The Neural-Image-QA \cite{malinowski2015ask} system uses VGG-Net image features \cite{simonyan2014very} and one-hot-encoded word representations are given as input to Long short term memory (LSTM) network for generating question features. Authors in \cite{antol2015vqa,Agrawal2017} have fused extracted image features (VGG-Net) and LSTM encoded question vector by element-wise multiplication. The $4096$-dimensional image features in \cite{ren2015exploring} are transformed into a vector (of same size as word embedding dimension). The modified and combined embeddings are given as input to LSTM for generating answer. In \cite{gao2016compact}, authors have proposed the fusion of multi-modal features through outer product (Bilinear pooling) as it provides multiplicative interaction (rich representation) between all elements of modalities. Bilinear pooling based fusion achieves superior performance, but seems to be a less efficient solution as a large number of parameters are needed for projection of outer product to obtain a joint representation of both modalities. However, later works in \cite{fukui2016multimodal,kim2016hadamard} have proposed Multimodal Compact Bilinear Pooling (MCB) and Multimodal Low-rank Bilinear (MLB) pooling, respectively for efficient use of bilinear pooling.

\subsection{Attended Feature Fusion}
\label{subsec:attFeatFuse}

Attention-based models \cite{shih2016look,yang2016stacked,kazemi2017show,lu2016hierarchical,xu2016ask} focus on the image region(s) that is (are) most relevant to the task (question). In VQA, attention models aim to interpret ``where to look" in the image for answering the question. Existing works have used attention in different ways. The attention can be on image \cite{yang2016stacked}, on question \cite{xu2016ask}, or on both (Co-attention) \cite{lu2016hierarchical}. For example, \cite{shih2016look} proposed a model that predicts the answer by selecting an image region which is most relevant to question text. 

A multi-step attention based method is proposed in \cite{yang2016stacked} that allows reasoning over fine-grained information asked in a question. Question embeddings used to generate attention distribution over image regions. The attention score obtained from weighted sum of image region embeddings are used as a visual feature for next step. The attention mechanism is used with outer product based fusion of image and question embeddings \cite{fukui2016multimodal}. Multimodal Factorized Bilinear (MFB) \cite{yu2017multi} pooling has been introduced to efficiently and effectively combine multi-modal features on top of low-rank bilinear pooling technique \cite{kim2016hadamard}. The usage of a stack of dense co-attention layers is proposed in \cite{nguyen2018improved}. Here, each word of a question interacts with each region proposal in an image and vice-versa. A combination of top-down and bottom-up attention models is proposed in \cite{anderson2018bottom}. The bottom-up model detects salient regions extracted using Faster-RCNN \cite{ren2015faster}, while the top-down mechanism uses task-specific context to predict attention score of the salient image regions.

A Question-Conditioned Graph (QCG) is processed for VQA in \cite{norcliffe2018learning}. Here, the objects proposed from faster-RCNN act as nodes and edges define the interaction between regions conditioned on question. For each node, a set of nodes is chosen from the neighborhood using strongest connection criterion. This leads to a question specific graph structure. Bilinear Attention Network (BAN) \cite{kim2018bilinear} fuses both the modalities by the interaction of each region proposal with each word of the question and uses residual connections to provide multiple attention glimpses. In Relation Network (RN)\cite{santoro2017simple}, every pair of object proposal embeddings are aggregated (summed up) and it is found that the resulting vector encodes the relationship between different regions thereby enabling compositional reasoning. In Question Type guided Attention (QTA) \cite{shi2018question}, semantics of question category are used with both bottom-up,top-down and residual features. A recurrent deep neural network with attention mechanism is proposed in \cite{noh2016training}, where each network is capable of predicting the answer. Dynamic Fusion With intra-and inter-modality Attention Flow (DFAF) \cite{gao2019dynamic} is a stacked network that uses inter-modality and intra-modality information for fusing features. Here, the use of average pooled features can dynamically change intra-modality information flow. The Multimodal Latent Interaction (MLIN) is proposed in \cite{gao2019multi} that realizes multi-modal reasoning through the process of summarization, interaction and aggregation. A generalized algorithm RAMEN is proposed in  \cite{shrestha2019answer} to deal with VQA datasets containing either only synthetic or real world images. 

This proposal uses top-down attention scores for fusing image and question embeddings. The answer space is decomposed into smaller sub-spaces based on specific question categories. A two stage hierarchical process is followed to predict answers (stage-2) based on predicted question category (stage-1). Our proposal is discussed next.

\section{Proposed Work}
\label{sec:propWork}

\begin{figure*}
\centerline{\includegraphics[width=1\textwidth,height=8cm,keepaspectratio]{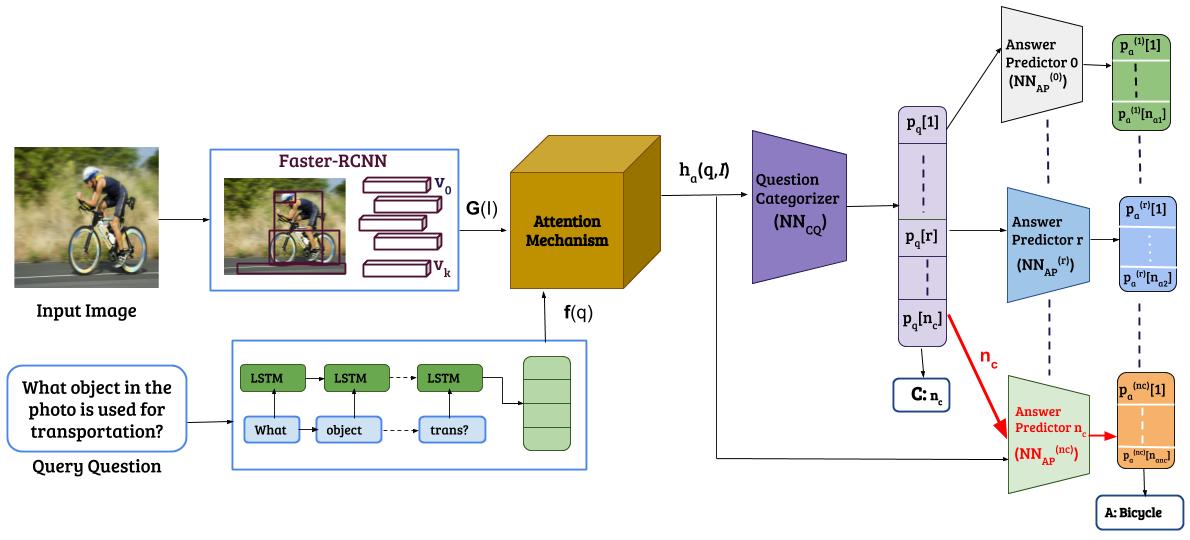}}
\caption{Illustrating the proposed approach. ResNet-101 features of regions proposed by faster-RCNN are extracted for visual representation. The question is encoded using a LSTM. Features of both modalities are fused by using region scores from a top-down attention model. The fused embedding is input to the \textbf{Question Categorizer} which selects one \textbf{Answer Predictor} (from multiple classifiers) to identify the output answer. For illustration, the $n_c^{th}$ category got highest score from category selection network. hence, the classifier corresponding to $n_c^{th}$ category will be active (shown in red) for final answer prediction.}
\label{fig:framework}
\end{figure*}

A visual question answering (VQA) system $\mathcal{S}_{VQA}$ aims at estimating the probabilities of answers $\mathbf{a}$ ($\mathbf{a} \in \mathbf{A}$) to an input (natural language) question $\mathbf{q}$ ($\mathbf{q} \in \mathbf{Q}$) about an image $I$ ($I \in \mathbf{I}$). Such a system is trained on the set of all images $\mathbf{I}$, set of questions $\mathbf{Q}$ associated with images and set of all answers $\mathbf{A}$. This is generally achieved by using representative vector space embeddings of questions ($\mathbf{f}(\mathbf{q})$) and images ($\mathbf{g}(I)$) computed using deep neural networks. The most probable answer $\mathbf{\hat{a}}$ is predicted by $\mathcal{S}_{VQA}$ as
	
\begin{equation}
\mathbf{\hat{a}} = \argmax_{\mathbf{a} \in \mathbf{A}} P(~\mathbf{a}~|~\mathcal{S}_{VQA}(~\mathbf{f}(\mathbf{q})~,~\mathbf{g}(I)~)~)
\label{eq:vqaSys}
\end{equation}

This proposal approaches VQA using a hierarchical architecture (Figure~\ref{fig:framework}) involving different answer prediction sub-systems corresponding to distinct question categories. This requires suitable deep networks for computing question and image features (vector space embeddings). These features of different modalities are fused using attention information. The process of feature extraction (Subsections~\ref{subsec:visFeatX} and~\ref{subsec:qFeatX}) and attention score guided feature fusion (Subsection~\ref{subsec:attMod}) are described next. 

\subsection{Visual Feature Extraction} 
\label{subsec:visFeatX}
The \emph{Visual Features} of images are extracted as embeddings by using a pre-trained deep network. Existing works \cite{anderson2018bottom,kim2018bilinear,shrestha2019answer} in VQA have mostly used \emph{Faster-RCNN} \cite{ren2015faster} for visual feature extraction. This model employs \emph{ResNet-101} \cite{he2016deep} as its base network and uses top-$k$ region proposals ($\mathbf{R}_{i}$; $i = 1,\ldots k$) for visual feature extraction. Let $\mathbf{v}_{i}$ ($\mathbf{v}_{i} \in \mathcal{R}^{d_{v}}$) be the ResNet-101 feature extracted from $\mathbf{R}_{i}$. The image $I$ is represented by the set of visual features $\mathbf{G}(I) = \{ \mathbf{v}_{i}; i = 1, \ldots k\}$. Experimental results have shown that a higher value of $k$ leads to a better representation at the expense of significantly higher computations. This proposal also uses the Faster-RCNN model with $k=36$ \cite{anderson2018bottom,shrestha2019answer}. This is followed by question feature extraction and is described next.

\subsection{Question Feature Extraction} 
\label{subsec:qFeatX}

The \emph{Question Features} are computed by using pre-trained deep netwroks. All questions are padded or truncated to obtain word sequences of a fixed length ($n_{w}$, say). The pre-trained GloVe embedding \cite{pennington2014glove} is used to convert a question $\mathbf{q}$ to an ordered sequence of word embeddings $\mathbf{E_{w}}(\mathbf{q}) = \{ \mathbf{ew}_{j}: \mathbf{ew}_{j} \in \mathcal{R}^{d_{w}}; j = 1, \ldots n_{w} \}$. This sequence of word embeddings are fed to a LSTM network $\mathbf{Q_{LSTM}}$ to generate the question embedding $\mathbf{f}(\mathbf{q})$. The $j^{th}$ hidden state embedding of $\mathbf{Q_{LSTM}}$ is obtained for each input word embedding $\mathbf{ew}_{j}$. The question embedding is obtained as the output of the final hidden state of $\mathbf{Q_{LSTM}}$ as $\mathbf{f}(\mathbf{q}) = \mathbf{Q_{LSTM}}( \mathbf{q} )$ ($\mathbf{f}(\mathbf{q}) \in \mathcal{R}^{d_q}$). 
The architecture of $\mathbf{Q_{LSTM}}$ is adopted from the LSTM network used in \cite{hochreiter1997long}.

The features extracted from visual (image) and text (question) modalities are fused using scores obtained from a top-down attention model. This attention mechanism is described next.

\subsection{Attention Mechanism}
\label{subsec:attMod}

\emph{Attention} plays a key role in fusing visual and question features. Attention guided feature fusion is adopted in several existing works (Sub-section~\ref{subsec:attFeatFuse}). Only a few among top-$k$ region proposals (identified during visual extraction) are relevant with respect to an input question $\mathbf{q}$. An attention network provides different scores to these region proposals using $\mathbf{f}(\mathbf{q})$ and $\mathbf{G}(I)$. Attention score guided feature fusion is performed to obtain the embedding $\mathbf{h_{a}}( \mathbf{q}, I)$. This process is described next.

\begin{figure}
\centerline{\includegraphics[width=\textwidth,height=8cm,keepaspectratio]{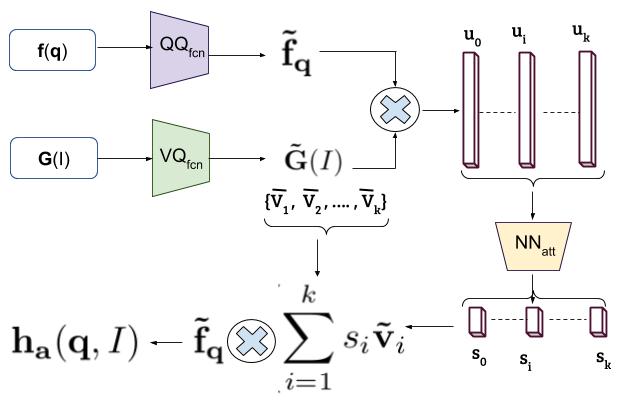}}
\caption{The functional block diagram of top-down attention network score guided fusion of visual and question features.}
\label{fig:attMod}
\end{figure}

The visual and question features are of different dimensions. Two fully connected networks $\mathbf{VQ_{fcn}}$ ($\mathbf{VQ_{fcn}}: \mathcal{R}^{d_v} \rightarrow \mathcal{R}^{d_f}$) and $\mathbf{QQ_{fcn}}$ ($\mathbf{QQ_{fcn}}: \mathcal{R}^{d_q} \rightarrow \mathcal{R}^{d_f}$) are used to map both visual and question features to vectors of size $d_{f}$. Both $\mathbf{VQ_{fcn}}$ and $\mathbf{QQ_{fcn}}$ are fully connected networks where the input and output layers are directly connected without any intermediate hidden layer. These two networks are used to map both visual and question embeddings to $\mathcal{R}^{d_f}$ as

\begin{eqnarray}
\mathbf{\tilde{v}}_{i} & = & \mathbf{VQ_{fcn}}( \mathbf{v}_{i} ) \\
\mathbf{\tilde{f}_{q}} & = & \mathbf{QQ_{fcn}}( \mathbf{f}(\mathbf{q}) ) 
\label{eq:vqProj}
\end{eqnarray}

These networks provide us with $\mathbf{\tilde{G}}(I) = \{ \mathbf{\tilde{v}}_{i}; i = 1, \ldots k\}$ and $\mathbf{\tilde{f}_{q}}$. Let $\mathbf{u}_{i} = \mathbf{\tilde{v}}_{i} \otimes \mathbf{\tilde{f}_{q}}$ be the element-wise product of $\mathbf{\tilde{v}}_{i}$ and $\mathbf{\tilde{f}_{q}}$. The vector $\mathbf{u}_{i}$ ($\mathbf{u}_{i} \in \mathcal{R}^{d_f}$) is input to the attention network $\mathbf{NN_{att}}$ to obtain the attention score $s_{i}$ corresponding to region proposal $\mathbf{R}_{i}$ ($i = 1, \ldots k$). The attention network $\mathbf{NN_{att}}$ is a fully connected network ($\mathbf{NN_{att}} : \mathcal{R}^{d_f} \rightarrow (0,1)$) that directly connects the input to a single-valued output without any intermediate hidden layer. The final attention score weighted feature fusion is performed to obtain $\mathbf{h_{a}}( \mathbf{q}, I )$ as

\begin{equation}
\mathbf{h_{a}}( \mathbf{q}, I ) = \mathbf{\tilde{f}_{q}} \otimes \left( \sum_{i=1}^{k} s_{i}\mathbf{\tilde{v}}_{i} \right)
\label{eq:attVec}
\end{equation}

\noindent where $\mathbf{h_{a}}( \mathbf{q}, I ) \in \mathcal{R}^{d_f}$. The process of attention score guided feature fusion is illustrated in Figure~\ref{fig:attMod}. The value of $\mathbf{h_{a}}( \mathbf{q}, I )$ depends on parameters of $\mathbf{Q_{LSTM}}$, $\mathbf{NN_{att}}$, $\mathbf{VQ_{fcn}}$ and $\mathbf{QQ_{fcn}}$. The parameters of these networks are tuned by minimizing the net loss (equation~\ref{eq:netLoss}) defined over the proposed hierarchical model CQ-VQA. The CQ-VQA model and the associated loss functions are discussed next.

\subsection{CQ-VQA: Learning the Model}

This work proposes a hierarchical model for visual question answering. This hierarchical model has two levels. At first level, the attention guided fused feature $\mathbf{h_{a}}(\mathbf{q},\mathbf{I})$ is used to classify the input question $\mathbf{q}$ into one of $n_{c}$ categories. Note that $n_{c}$ depends on the dataset under consideration. For example TDIUC (Section~\ref{subsec:DS}) has $n_c = 12$ question categories. The first level uses a single hidden layer feedforward network $\mathbf{NN_{CQ}}$ ($\mathbf{NN_{CQ}} : \mathcal{R}^{d_q} \rightarrow (0,1)^{n_c}$) to perform the task of question classification. 

Let $\mathbf{t_q}$ be the one-hot-encoded target vector representing the ground truth question category $\mathbf{q_c}$. Let $\mathbf{p_q}$ be the output of $\mathbf{NN_{CQ}}$. The question classification loss is defined as

\begin{equation}
\displaystyle \mathcal{L}_{Q}(\mathbf{q},I,\mathbf{q_c}) = {-}\sum_{r=1}^{n_c} \mathbf{t_q}[r]\log(\mathbf{p_q}[r])
\label{eq:lossQ}
\end{equation}

The second level of the hierarchy in CQ-VQA predicts the answers based on input question and image. Generally, the answer search space is large. This proposal decomposes the answer set $\mathbf{A}$ into $n_c$ subsets $\mathbf{A}_{r}$ according to the question categories. Thus, $\mathbf{A}_{r} \subset \mathbf{A}$ ($r=1, \ldots n_c$) and $\displaystyle \cup_{r=1}^{n_c} \mathbf{A}_{r} = \mathbf{A}$. The question classification network $\mathbf{NN_{CQ}}$ acts as a switch for selecting one of $n_c$ answer prediction sub-systems. Each answer prediction sub-system is a VQA system capable of predicting one from a subset of $\mathbf{A}$ based on the question category. We believe that this answer search space decomposition makes the task of VQA easier by reducing the number of outputs for each answer predictor. For example, questions of the form ``Is there a bird in the image?'' are of the binary answer (yes/no) category and the corresponding answer prediction sub-system has only two outputs. Similarly questions asking for ``What color is the bird?'' has only a small number of answers (colors) to choose from a small subset of $\mathbf{A}$. 

Let $n_{a}^{(r)}$ be the number of possible answers for the $r^{th}$ question category. The target answer $\mathbf{a}$ is one-hot-encoded through the $n_{a}^{(r)}$ dimensional vector $\mathbf{t_{a}^{(r)}}$. The attention guided fused feature $\mathbf{h_{a}}( \mathbf{q}, I )$ is input to the $r^{th}$ answer prediction sub-system $\mathbf{NN_{AP}^{(r)}}$ for predicting the answer probability vector $\mathbf{p_a}^{(r)}$ ($\mathbf{p_a}^{(r)} \in (0,1)^{n_{a}^{(r)}}$). The answer prediction networks are fully connected networks with single hidden layer. The loss $\mathcal{L}_{A}^{(r)}$ for training $\mathbf{NN_{AP}^{(r)}}$ is defined as

\begin{equation}
\displaystyle \mathcal{L}_{A}^{(r)}(\mathbf{q},I,\mathbf{a}) = - \sum_{j=1}^{ n_{a}^{(r)} } \mathbf{t_a}^{(r)}[j] \log( \mathbf{p_a}^{(r)}[j] )
\label{eq:nnarLoss}
\end{equation}

The net loss at the second level is defined as

\begin{eqnarray}
\displaystyle \mathcal{L}_{AA}(\mathbf{q},I,\mathbf{a}) & = & \sum_{r=1}^{ n_c } \delta[r - \rho] \mathcal{L}_{A}^{(r)}(\mathbf{q},I,\mathbf{a}) \\
\displaystyle \rho & = & \argmax_{l=1, \ldots n_c} \mathbf{p_q}[l]
\label{eq:aaLoss}
\end{eqnarray}

\noindent where $\delta[i-j]$ is the Kronecker delta function. The overall loss of CQ-VQA for input question $\mathbf{q}$, its category $\mathbf{q_c}$, associated image $I$ and ground-truth answer $\mathbf{a}$ is given by

\begin{equation}
\mathcal{L}_{CQVQA}(\mathbf{q}, \mathbf{q_c}, I , \mathbf{a}) = \mathcal{L}_{Q}(\mathbf{q},I,\mathbf{q_c}) + \mathcal{L}_{AA}(\mathbf{q},I,\mathbf{a})
\label{eq:netLoss}
\end{equation}

This proposal minimizes the loss $\mathcal{L}_{CQVQA}(\mathbf{q}, \mathbf{q_c}, I , \mathbf{a})$ for all question-image-answer combinations $(\mathbf{q}, I, \mathbf{a}) \in \mathbf{Q} \times \mathbf{I} \times \mathbf{A}$. The gradients computed by using this net loss (equation~\ref{eq:netLoss}) are back-propagated for end-to-end training of $\mathbf{Q_{LSTM}}$, $\mathbf{VQ_{fcn}}$, $\mathbf{QQ_{fcn}}$, $\mathbf{NN_{att}}$, $\mathbf{NN_{CQ}}$ and $\mathbf{NN_{AP}^{(r)}}$ ($r = 1, \ldots n_c$). 

\section{Experiments}
\label{sec:expRes}

This section briefly discusses the dataset, evaluation metrics, and implementation details.

\subsection{Dataset: TDIUC}
\label{subsec:DS}

\begin{figure}[h!]
\centerline{\includegraphics[width=0.75\textwidth]{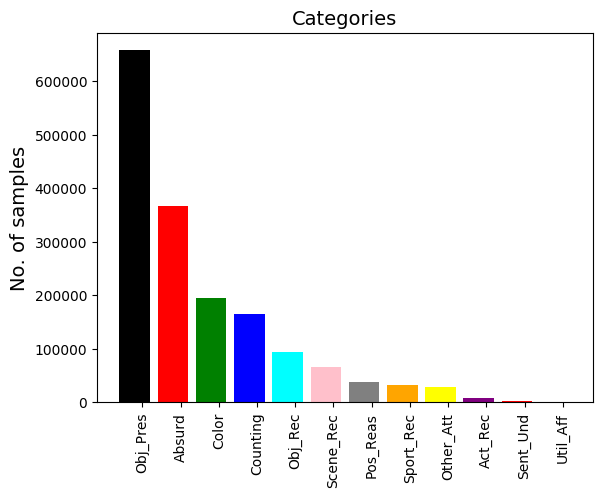}}
\caption{Distribution of $12$ Categories of TDIUC Questions \cite{kafle2017analysis}.}
\label{fig:TDIUC}
\end{figure}

Task-Directed Image Understanding Challenge (TDIUC) \cite{kafle2017analysis} is the largest available VQA dataset of real images. TDIUC consists of $16,54,167$ open-ended questions of $12$ categories associated with $1,67,437$ images. Questions in TDIUC are acquired from the following three sources. First, questions imported from existing datasets; second, questions generated from image annotations; and third, questions generated through manual annotations. Figure~\ref{fig:TDIUC} shows the category-wise sample distribution of questions. The largest number of questions (approximately $~6.5$ million) are in the `Object Presence' (with Yes/No answers) category. On the other hand, the least number of questions (only $521$) lie in the `Utility Affordance' category. The `Absurd' is an exceptional category consisting of questions having no semantic relation with associated image input. Such questions have a single answer and that is `Does-Not-Apply' \cite{kafle2017analysis}. Researchers have observed the phenomenon of model bias towards language priors. The introduction of the `Absurd' forces the model to learn proper relations between question(s) and visual contents of image(s).

\subsection{Evaluation Metrics}
\label{subsec:evalMet}

This proposal employs three commonly used evaluation metrics for the VQA task. These are \textit{Overall accuracy}, \textit{Arithmetic-Mean Per Type (MPT)} and \textit{Harmonic-Mean Per Type (MPT)}. 
The \textit{Overall accuracy} is the ratio of the number of correctly answered questions to the total number of questions. VQA datasets are highly imbalanced as a few question categories are more frequent than others. \textit{Overall accuracy} is not a good evaluation metric for such cases. The other two metrics \textit{Arithmetic-Mean Per Type (MPT)} and \textit{Harmonic-Mean Per Type (MPT)} \cite{kafle2017analysis} are generally used to achieve unbiased evaluation. \textit{Arithmetic-MPT} computes the arithmetic mean of the individual accuracies of each question category. This evaluation metric assigns uniform weight to each question category. \textit{Harmonic-MPT} reports the harmonic mean of individual question category accuracies. Unlike Arithmetic-MPT, the Harmonic-MPT measures the ability of a model to have a high score across all question categories.

    \begin{table}[t]
    \centering
		\addtolength{\tabcolsep}{15pt}   
		\caption{Comparing \textit{Overall Accurary} of CQ-VQA and other state-of-art models. CQ-VQA outperforms all models except MLIN. The higher accuracy of MLIN (marked with $\star$) can be attributed to its usage of top $100$ region proposals for visual feature extraction, while all other models (including CQ-VQA) have used only top-$36$ regions.}
		\begin{tabular}{l|c}
			\hline
			\textbf{Model} & \textbf{Overall Accuracy}\\
			\hline 
			
			\textbf{BTUP \cite{anderson2018bottom}} & 82.91\\
			\textbf{QCG \cite{norcliffe2018learning}}  & 82.05\\
			\textbf{BAN \cite{kim2018bilinear}}  & 84.81\\
			\textbf{RN \cite{santoro2017simple}} & 84.61\\
			\textbf{DFAF \cite{gao2019dynamic}} & 85.55\\
			\textbf{RAMEN \cite{shrestha2019answer}}& 86.86\\
			\textbf{MLIN$\star$ \cite{gao2019multi}}& \textbf{87.60} \\
			\hline
			\textbf{CQ-VQA}& 87.52\\
			\hline
			
		\end{tabular}
		\label{tab:overallPerf}
	\end{table}
	
			\begin{table*}[h!]
			\centering
			\addtolength{\tabcolsep}{8pt}    
			\caption{Category-wise performance comparison with state-of-the-art methods on TDIUC dataset}
			\begin{tabular}{l|c|c|c|c|c}
				\hline
				\textbf{Question Type} &\textbf{NMN} & \textbf{RAU} & \textbf{MCB} & \textbf{QTA} & \textbf{CQ-VQA}\\
				&\textbf{\cite{DBLP:journals/corr/AndreasRDK15}}&\textbf{\cite{noh2016training}} & \textbf{\cite{gao2016compact}} & \textbf{\cite{shi2018question}} &    \\
				\hline
				{Scene Recognition}& 91.88 & 93.96 & 93.06 & 93.80 & \textbf{94.05}\\
				{Sport Recognition}& 89.99 & 93.47 & 92.77 & \textbf{95.55} &   95.39\\
			    {Color Attributes}&54.91 & 66.86 & 68.54 & 60.16 &    \textbf{73.35}\\
				{Other Attributes}&47.66 & 56.49 & 56.72 & 54.36 &  \textbf{59.24}\\
				{Activity Recognition}&44.26 & 51.60 & 52.35 & 60.10 &   \textbf{61.19}\\
				{Positional Reasoning}&27.92 & 35.26 & 35.40 & 34.71 &    \textbf{40.40}\\
				{Object Recognition}&82.02 & 86.11 & 85.54 & 86.98 &   \textbf{88.13}\\
				{Absurd}&87.51 & 96.08 & 84.82 & \textbf{100.0} &  \textbf{100.0}\\
				{Utility \& Affordance}& 25.15 & 31.58 & \textbf{35.09} & 31.48 &  34.50\\
				{Object Presence} & 92.50& 94.38 & 93.64 & 94.55   & \textbf{95.41}\\
				{Counting} &49.21& 48.43 & 51.01 & 53.25 & \textbf{56.78}\\
				{Sentiment Und.} &58.04& 60.09 & 66.25 & 64.38 &   \textbf{66.56}\\
				\hline
				{Overall Accuracy}&79.56 & 84.26 & 81.86 & 85.03 &  \textbf{87.52}\\
				{Arithmetic-MPT} &62.59& 67.81 & 67.90 & 69.11 &   \textbf{72.08}\\
				{Harmonic-MPT}&51.87 & 59.00 & 60.47 & 60.08 &   \textbf{64.45}\\
				\hline
			\end{tabular}\label{tab:classwiseRes}
		\end{table*}

\subsection{Implementation Details}
\label{subsec:impDet}

The top-$36$ ($k = 36$) region proposals of ResNet-101 are used to compute $d_v = 2048$ dimensional visual feature vectors. The length of each question is set to $n_w = 14$ words. Questions with more than $14$ words are truncated and lesser than that are padded with zero embedding vectors. The pre-trained GloVe network is used to generate word embeddings of size $d_w = 300$. A sequence of these word embeddings are input to a LSTM ($\mathbf{Q_{LSTM}}$, Subsection~\ref{subsec:qFeatX}) for question embedding generation. The sizes of hidden and output layer of $\mathbf{Q_{LSTM}}$ are both set to $1024$. Thus, the question embeddings are of size $d_q=1024$. For attention module, both visual features $\mathbf{v}_{i}$ ($i=1,\ldots k$) and question features $\mathbf{f}(\mathbf{q})$) are projected to $1024$ dimensional space. These $d_f=1024$ dimensional vectors are further processed for attention score weighted feature fusion (Subsection~\ref{subsec:attMod}).

The TDIUC dataset contains $12$ question categories. Thus, the question categorization network $\mathbf{NN_{CQ}}$ predicts the vector $\mathbf{p_q}$ of size $n_c = 12$. Accordingly, one network $\mathbf{NN_{AP}}^{(r)}$ (from $n_c=12$) is selected to predict the answer $\mathbf{a}$ using $d_f = 1024$ dimensional fused feature $\mathbf{h_{a}}(\mathbf{q}, I)$. The complete model is trained in an end-to-end manner for $17$ epochs with a batch size of $512$. 
The Adamax optimizer \cite{kingma2014adam} is used with a decaying step learning rate. The initial learning rate is set to $0.002.$ with a decay factor of $0.1$ after $5$ epochs.

\section{Results \& Discussion}
\label{sec:resDisc}

This section discusses a comparative performance analysis of CQ-VQA and other state-of-art methods (Subsection~\ref{subsec:perfComp}). An ablation analysis is performed to understand the effectiveness of the proposed model (CQ-VQA). The results of this analysis are reported in Subsection~\ref{subsec:abAnls}.

\subsection{Comparison with State-of-Art Methods}
\label{subsec:perfComp}

The performances of different VQA methods are compared under two settings. The first setting compares overall accuracy of all models. There are VQA approaches for which, we do not have access to question category-wise results (not available in literature). Such models are primarily compared in the first setting. Table \ref{tab:overallPerf} presents the accuracy of different methods. 
Results shown in bold represents the best performance among all models. The overall accuracy obtained by MLIN \cite{gao2019multi} and proposed CQ-VQA is comparable. However, it is noteworthy to mention that MLIN (marked with $\star$) has used top-$100$ regions to extract visual features, while all other models (including CQ-VQA) have used only top-$36$ regions. As discussed earlier (Subsection~\ref{subsec:visFeatX}), a higher number of region proposals ($k$) leads to improved performance at the cost of significantly higher computation.

Question category-wise VQA performance of models are compared in the second setting. Here, only those VQA approaches are considered for which such results are available in the literature. Table~\ref{tab:classwiseRes} shows the question category-wise accuracy of all methods compared in the study. The last three rows represent the comparisons of the three evaluation metrics for all VQA models under consideration. 

Table~\ref{tab:classwiseRes} shows that CQ-VQA is the best performer on all three evaluation metrics. Further, at the category-wise performance, CQ-VQA is the best performer for $10$ out of $12$ classes. In the other two categories, \textit{sport recognition} and \textit{utility and affordance}, CQ-VQA is the second-best performer. For some question categories, a significant performance improvement is obtained by CQ-VQA. For example, CQ-VQA obtains an improvement of $~7\%$ and $~14\%$ for `color' and  `Positional Reasoning' categories, respectively.

\subsection{Ablation Studies}
\label{subsec:abAnls}

The proposed approach leverages on question categories to solve the VQA problem. An ablation analysis is conducted to show the efficacy of the hierarchical approach of CQ-VQA. In this analysis, a baseline model is constructed by removing \textit{Question Categorization} and \textit{Answer Predictor} components of CQ-VQA. However, the baseline uses the same set of attended and fused features as CQ-VQA. Results shown in Table-\ref{tab:ablation} shows a relative improvement of $1.45\%$ by CQ-VQA in terms of overall accuracy. CQ-VQA shows improved performance on the other two evaluation metrics as well.

The effect of language bias prior is commonly observed in VQA. This is analysed next. In TDIUC dataset, the `Absurd' category is introduced to test the effect of language prior biases in model performance. Our experiment compares the performance of CQ-VQA under two settings -- with and without the `Absurd' category of questions. Table~\ref{tab:woabs} shows a significant drop in performance indicating that CQ-VQA is also affected by language prior biases.

    \begin{table}[H]
        \centering
        \caption{Ablation analysis: Effect of removing hierarchy from CQ-VQA}
        \addtolength{\tabcolsep}{10pt}    
        \begin{tabular}{l|c|c}
    
            \hline
            \textbf{Metrics} & \textbf{Baseline}& \textbf{CQ-VQA}\\
            \hline
            \textbf{Overall Accuracy} & 86.26 &\textbf{87.52}\\
            \textbf{Arithmetic-MPT} & 70.71 &\textbf{72.08}\\
            \textbf{Harmonic-MPT} & {63.37} & \textbf{64.45}\\
            \hline
        \end{tabular}
        
        \label{tab:ablation}
    \end{table}
    
    \begin{table}[H]
        \centering
        \caption{Ablation analysis: Performance of CQ-VQA on the data (except Absurd category samples) trained using with and without `Absurd' Category samples}
        \addtolength{\tabcolsep}{7pt}    
        \begin{tabular}{l|c|c|c}
        \hline
         \multicolumn{4}{c}{\textbf{Without Absurd}}\\
            \hline
            \textbf{Metrics} & \textbf{MCB} &\textbf{QTA} & \textbf{CQ-VQA}\\
            \hline
            \textbf{Overall Accuracy} & 78.06 & 80.95 &\textbf{83.46}\\
            \textbf{Arithmetic-MPT} & 66.07 & 66.88 &\textbf{68.69}\\
            \textbf{Harmonic-MPT} &55.43 & 58.82 & \textbf{61.44}\\
            \hline
        \end{tabular}
        \label{tab:woabs}
    \end{table}

\section{Conclusion \& Future Work}
\label{sec:conc}
In this work, a novel hierarchical end-to-end model CQ-VQA is presented for the VQA task. CQ-VQA leverages over question categorization to reduce the potential answer search space. Empirical results on the TDIUC dataset indicate that the performance of CQ-VQA is competitive with respect to state-of-art VQA methods. 

The performance of the proposed model can be further enhanced by using better feature extractor proposals, attention mechanisms and more complex question/answer prediction networks. Also, a challenge remains for datasets where question category ground-truth are not available. We plan to work in that direction as a natural extension of the present proposal.


\bibliography{references.bib} 

\newpage

\end{document}